\documentclass[letterpaper, 10 pt, conference]{ieeeconf}  % Comment this line out if you need a4paper

\IEEEoverridecommandlockouts                              % This command is only needed if 
\usepackage{graphicx}

\title{\LARGE \bf
Video Transformer for Remote Identity Document Hologram Detection}

\author{Joris Voerman$^{1}$, Nicolas Sidere$^{1}$ and Jean-Christophe Burie$^{1}$% <-this % stops a space
\thanks{$^{1}$All authors are from La Rochelle University, 17031 La Rochelle, France {\tt\small \{joris.voerman,nicolas.sidere, jean-christophe.burie\}@univ-lr.fr}}%
}

\begin{document}

\maketitle
\thispagestyle{empty}
\pagestyle{empty}

%%%%%%%%%%%%%%%%%%%%%%%%%%%%%%%%%%%%%%%%%%%%%%%%%%%%%%%%%%%%%%%%%%%%%%%%%%%%%%%%
\begin{abstract} %Rework Intro to include RIDVS
%Hologram analysis remains a significant challenge in remote Identity Document (ID) verification, particularly for detecting forgeries. Despite its importance, there is currently no suitable solution in the State-of-the-Art (SotA) outside of specially designed hardware for hologram capture. Yet, holograms are a crucial security feature: they are highly resistant to falsification and can be observed in standard video captures, unlike other features such as infrared properties.
Remote identity authentification using Identification Documents has been a major challenge for several years. DeepFakes advent and the development of AI-guided tools helps fraudsters creating counterfeit ID Documents. Ensuring the authenticity of ID Documents has become a real clue in the seurization of remote authentification. This need is all the more pressing given the increasing digitization of administrative and transactional processes. To ensure widespread accessibility, the system should rely solely on video captured via mobile devices. In this specific context, confirming the authenticity of ID is a real challenge as many security features needs specific device like infrared sensor for instance. Among underutilized but promising security features, holographic printings hold a special place. Difficult to counterfeit, they produce distinctive visual effects according enlightment, making them both detectable in a video captured by a smartphone camera and difficult to imitate. 

%In this article, we propose an approach based on a video transformer for detecting holograms in simple videos captured by smartphones. Our method builds on a robust model previously validated in a related research domain. We demonstrate that it outperforms existing SotA methods, achieving near-perfect accuracy even when trained on medium- to small-sized datasets.
%This study includes several experiments that evaluate the model adaptation to frugality, both for training samples and computational resources.
In this paper, we propose a Remote Identity Document Verification System (RIDVS) and an approach based on a video transformer for detecting holograms in simple videos captured by smartphones. Our system is designed for a smartphone-based capture process, followed by a server-side verification. The hologram detection method builds on a robust model previously validated in a related research domain. We demonstrate that it outperforms existing SotA methods, achieving near-perfect accuracy even when trained on medium- to small-sized datasets. In particular, we report improvements of +26.86\% in Recall and +17.93\% in accuracy over the best MIDV-Holo baseline.
This study includes several experiments that evaluate the model adaptation to frugality, both for training samples and computational resources. %potentiellement retirer durant le degraissage

\end{abstract}

%%%%%%%%%%%%%%%%%%%%%%%%%%%%%%%%%%%%%%%%%%%%%%%%%%%%%%%%%%%%%%%%%%%%%%%%%%%%%%%%
\section{INTRODUCTION}\label{sec:intro} %Rework Intro to include RIDVS
This work is conducted within the framework of developing advanced solutions for remote ID verification via smartphone applications, primarily for banking and insurance services. The demand for such solutions has grown considerably in recent years, a trend further accelerated by the COVID-19 crisis and its aftermath. Ensuring the authenticity of an ID in remote settings is crucial for preventing identity theft and other forms of fraud.

However, verifying ID documents through user-recorded videos presents substantial challenges. Many traditional security features, especially those designed to prevent counterfeiting, cannot be validated under such conditions, which increases the risk of fraud. Among the few usable features in this context, the hologram stands out. It is extremely difficult to forge without professional equipment and cannot be easily simulated, making it a key component in the fight against document fraud. Despite its potential, hologram detection remains underexplored in the document verification community. This is largely due to the technical challenges it entails, as reliable detection requires video analysis and 2D reconstruction.

The situation is further complicated by the lack of available training data. IDs are highly sensitive and protected; therefore, collecting even a small dataset with just a few samples is difficult. To our knowledge, only one public dataset is currently available for research purposes, and it contains only mock documents that are not even directly inspired by any real country. Adding to the challenges posed by this context is the issue of limited computing resources. Ideally, solutions deployed in such environments must be both fast and cost-effective, while still maintaining the best possible performance.

%The main two contributions are the proposal of a lighter transformer model for Hologram detection and an extensive evaluation, which includes MIDV-Holo various challenges and training/computing resources' frugality.

The main contributions of this paper are: (1) the proposal of an overall system for remote ID verification; (2) a lighter transformer model for hologram detection named ViTransHolo; (3) an analysis of the model behavior across all MIDV-Holo fraud scenarios; and (4) an extensive evaluation of training and computational frugality.

\section{REMOTE IDENTITY DOCUMENT VERIFICATION SYSTEM}\label{sec:system}

\begin{figure*}[tbhp]
\centering
\framebox{\parbox{7in}{
\includegraphics[width=\textwidth]{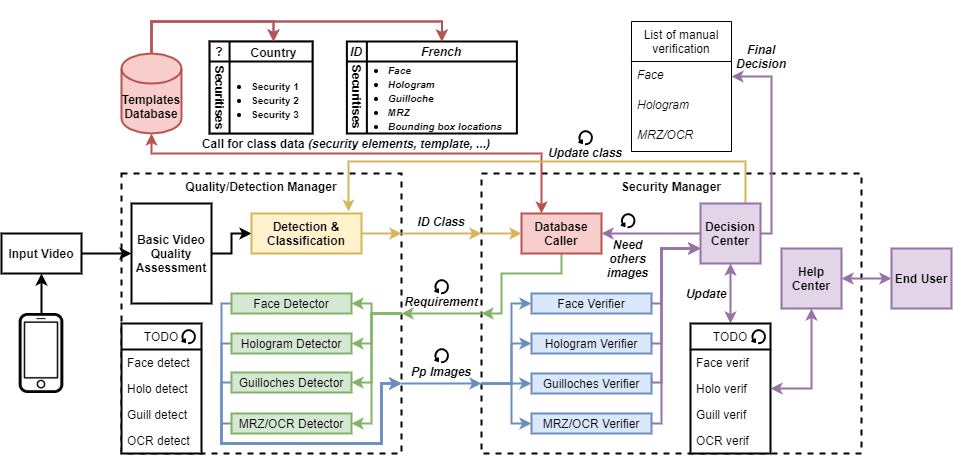}
}}
\caption{Overall diagram of the proposed system for remote ID authentication. The colors correspond to specifics sub-systems : yellow for classification, red for the database, green for detectors, blue for verifiers and purple for decision modules} \label{img:overall_Sys}
\end{figure*}

%Introduction du système sur la base du graphique général
Our proposal for a remote identity document verification system follows the overall diagram of Fig.~\ref{img:overall_Sys}. It consists of five main modules.

The first one is a document \textbf{Detection \& Classification} module used to identify the document and adapt the security element verification process to its template. This module is mandatory to include the essential prior knowledge for the subsequent steps. 
To this end, it's followed by a \textbf{Database} module that stores prior knowledge and templates of each known IDs. It uses the ID class as input and feed Detectors and Verifiers to optimise their process to the current situation.
The first step of the verification process is the detection modules (simplify as \textbf{Detectors}). It filters the image and ensures that the video capture is of sufficient quality for the corresponding verifier.
\textbf{Verifiers} exploit detected images to verify the conformity of the detected security elements with the expected template given by the database.
The final module is a \textbf{Decision Center}. It coordinates processing and updates the status of the overall verification. It could do an ID class correction (if another template seems to be more suitable) or trigger a new detection process if the verification fails or communicate with the \textbf{End User UI} to call extensive action (like redo the capture). It also generates the final decision. 
In addition to the main modules, the system includes a basic detector for overexposure and blurriness (just after the video capture) to trigger a retake as soon as possible in the event of unusable video. This could also happen if a document fails to be detected or classified.

Overall, the system is designed to keep as many components as possible on the device; ideally, the entire \textbf{Quality/Detection Manager} would be handled by the device, whilst the \textbf{Security Manager} would be located on the server. However, not all \textbf{Detectors} are necessarily suited to such an architecture, and some of them will, in all likelihood, require the greater computational resources of the server-side. Hologram detection, which requires extensive video processing, is a good example of this.

%Lien vers les autres travaux déjà réalisés dans le cadre du projet précédent
The classification module and the Guilloches verifier have already been addressed in previous studies. The classification is operated by a prototypical network classifier, which specialises in the use of a small number of templates to perform quick few-shot learning classification with good precision \cite{classif_module}. The Guilloches verification uses a Convolutional Network driven by band-pass filtering \cite{guilloches_module}.

%Lien avec la suite sur les hologrammes
To pursue the presentation of the remote identity document verification system, we will introduce the last finished module: the hologram detector. To this end, we will introduce the hologram detection State of the Art, the architecture and then demonstrate its efficiency in our context.

\section{RELATED WORKS}\label{sec:RW}

\begin{figure*}[tbhp]
\centering
\framebox{\parbox{7in}{
\includegraphics[width=\textwidth]{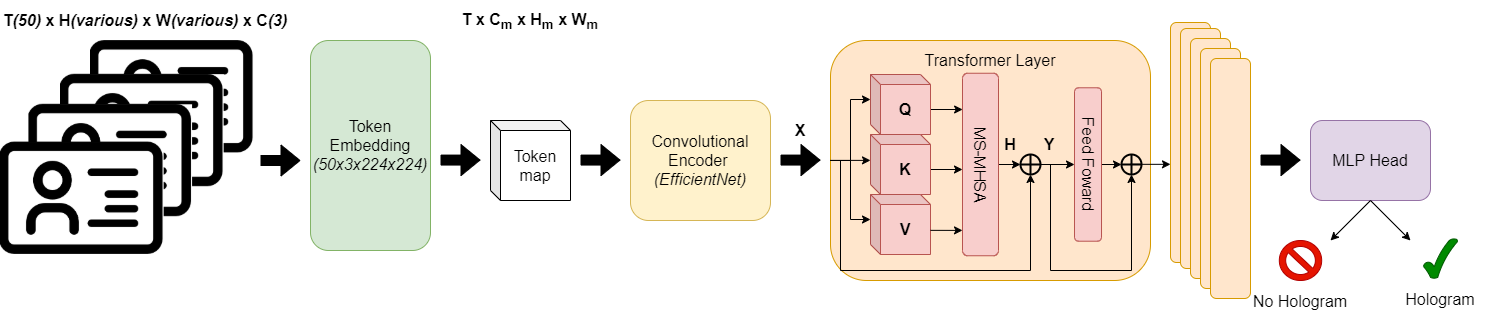}
}}
\caption{ViTransHolo model architecture. It begin with a embedding process to format input video, then an encoder (EfficientNet), followed by six transformer layers (using VitransPAD MS-MHSA mechanism) and a Multi-Layer Perceptron to provided final classification} \label{img:vitransholo}
\end{figure*}

%JCB: Reformulation pour gagner de la place  
%Several studies have addressed the problem of ID hologram detection and reconstruction, although the field remains relatively underexplored. One of the most notable contributions is the MIDV-Holo dataset, along with its associated baseline method proposed in \cite{midvholo}. This dataset serves as a benchmark in our study (see Section~\ref{subsec:dataset}). The baseline approach relies on pixel-wise classification using hue and saturation features extracted from video frames. While it achieves reasonably good results and partially reconstructs the hologram, it remains highly sensitive to noise, such as overexposure.
ID hologram detection and reconstruction remains a relatively underexplored field. Its most notable contribution is the MIDV-Holo dataset and associated baseline \cite{midvholo}, which serves as our benchmark (Section~\ref{subsec:dataset}). The baseline relies on pixel-wise classification from hue and saturation features, yielding reasonable results but proving sensitive to noise such as overexposure.
Building on the same dataset, Pouliquen et al. \cite{pouliquen2024weakly} introduced a weakly supervised classification model based on a triplet loss architecture. Their method is tailored to situations with limited annotated data and shows a modest improvement over the MIDV-Holo baseline, though the performance gain remains limited.

Outside of the MIDV-Holo dataset, several other approaches have been explored. Ay %& al.
\cite{ay2022open} proposed a method using Generative Adversarial Networks (GANs), but its limited scope and lack of comparability with standard benchmarks make it difficult to evaluate in the same context. Another noteworthy effort is the work of Chapel %et al.
\cite{chapel2023authentication}, which employs Local Binary Pattern (LBP) comparisons to detect holograms. Kada %et al.
\cite{kada2022hologram} also proposed a method that, like the MIDV-Holo baseline, uses handcrafted features based on hue and saturation. Neither Chapel %et al.
nor Kada %et al. 
report results on the MIDV-Holo dataset, which prevents direct quantitative comparison; we therefore focus our evaluation on the two directly comparable methods \cite{midvholo, pouliquen2024weakly}.

Several other studies, such as Soukup %et al.
\cite{soukup2017mobile}, rely on specialised hardware for hologram capture, rendering them unsuitable for smartphone-based applications. Similarly, multiple studies use hyperspectral images to detect holograms and forgeries \cite{mukundan2022portable}. Such methods are not applicable due to the impossibility of capturing hyperspectral images with a basic smartphone. Similarly, there is a significant body of work focused on biological or microscopic holography (e.g., \cite{zeng2021deep}), which lies outside the scope of ID document analysis.

In summary, current research in this field is limited, with only one publicly available dataset and few comparable approaches. This highlights the need for more robust and generalisable solutions for hologram detection in remote ID verification using videos captured on consumer smartphones.

\section{TRANSFORMER-BASED ARCHITECTURE FOR HOLOGRAM DETECTION}\label{sec:method}

%Pourquoi essayer un transformer
Since the early 2020s, transformer architectures have demonstrated impressive performance across a wide range of computer vision tasks, including video analysis. Notable examples include applications in remote sensing video detection \cite{jiao2023transformer}, object tracking \cite{xie2023videotrack}, 3D reconstruction \cite{wang2021multi}, and deepfake, forgery or presentation attack detection \cite{ming2022vitranspad}.

%Quel choix de base et pourquoi
In this work, we build upon a proven model from a related domain: ViTransPAD \cite{ming2022vitranspad}, a video transformer originally designed for face presentation attack detection. This model provides a solid foundation due to its demonstrated balance between accuracy and computational efficiency. Our adapted version, tailored for hologram detection, is referred to as ViTransHolo, named in tribute to the original architecture. An illustration of the modified architecture is provided in Fig.~\ref{img:vitransholo}. The implementation was developed using PyTorch.

%explication du processus de l'input jusqu'a la sorti en precisant les choix d'implémentation
\subsection{Model Pipeline}
Our approach processes input videos by selecting 50 equally spaced frames. This value was chosen to provide sufficient temporal coverage while remaining computationally tractable; a sensitivity analysis on the number of frames is left for future work. Each frame undergoes preprocessing (see \textbf{Token Embedding} block in Fig.~\ref{img:vitransholo}), where a document tracking algorithm crops the ID from the background and resizes it to 224×224 pixels to match the input size requirements of the model. The frames are then encoded into a PyTorch tensor with shape (Time$\times$Channel$\times$Height$\times$Width). 

For feature extraction, we use EfficientNet \cite{tan2019efficientnet} as \textbf{Convolutional Encoder}. It provides a good trade-off between computational cost and representational power, and is comparable to Vision Transformers (ViT) \cite{ViT} in our settings.

The extracted features are passed to a stack of six \textbf{Transformer Layers}, each enhanced with Multi-Scale Multi-Head Self-Attention (MS-MHSA). This MS-MHSA mechanism, introduced in ViTransPAD \cite{ming2022vitranspad}, improves the model’s ability to capture spatial-temporal dependencies by employing a pyramid-like structure that scales attention heads to generate multi-scale feature representations. For implementation details, we refer the reader to the original publication.
Each attention layer is followed by a \textbf{Feed-Forward} block, which consists of two sequential modules, each comprising a convolutional layer followed by a ReLU activation. This block is integrated within the standard transformer architecture.

Frame-level classification is performed using a Multi-Layer Perceptron (MLP) at the end of the model. 
%JCB: (pour raccourcir)
%The final video-level prediction is then obtained by averaging the individual frame-level outputs, effectively summarising the model’s decision across the entire video sequence.
Video-level predictions are derived by averaging frame-level outputs, effectively summarising the model's decision over the full sequence. 
%
%The resulting architecture can be viewed as a lightweight adaptation of ViTransPAD, specifically optimized for our hologram detection context. 
%JCB: (pour raccourcir)
%This streamlined architecture requires fewer training samples, consumes fewer computational resources, and offers faster inference times, making it well-suited for real-world deployment scenarios.
This streamlined design reduces training sample requirements and computational cost, making it practical for real-world deployment.
%
%I move it each time to be sur it is on the second page when I speak of it
%\begin{figure*}[tbhp]
%\includegraphics[width=\textwidth]{Images/ViTransHolo.png}
%\caption{ViTransHolo model architecture. It begin with a embedding process to format input video, then an encoder (EfficientNet), followed by six transformer layers (using VitransPAD MS-MHSA mechanism) and a Multi-Layer Perceptron to provided final classification} \label{img:vitransholo}
%\end{figure*}

%Some parameters details
\subsection{Training Details}
The model is trained using binary cross-entropy loss, with the Adam optimiser and a cosine annealing warmup-restart scheduler to stabilise and accelerate convergence. All experiments were performed on a high-performance server equipped with NVIDIA A40 GPUs.

\section{EXPERIMENTS}\label{sec:EXP}
\subsection{Dataset}\label{subsec:dataset}
For our experiments, we use two different datasets:
%MIDV-Holo and IML3.%IML3 n'est pas publiquement disponible

\subsubsection{MIDV-Holo Dataset}
MIDV-Holo \cite{midvholo} is a publicly available dataset composed of mock identity documents (ID cards and passports) from a fictional country. These synthetic documents contain holograms as part of their security features and are designed specifically to evaluate hologram detection and recognition methods. The dataset contains two main types of documents, ID cards and passports, each with 10 distinct templates. For each template, five variants are generated using fictitious personal data, resulting in 50 ID cards and 50 passports. Each of these mock documents is recorded under three different lighting conditions, leading to a total of 150 "original" videos per document type, or 300 videos in total, all labeled as "True" samples.

To simulate fraud scenarios, MIDV-Holo also includes "False" samples created using the same 100 mock documents. These are categorised into four types: (1) No hologram, (2) Photocopied hologram, (3) Pseudo-hologram (an artificial reflective surface), (4) Photo replacement, where the original photo (often overlaid with a hologram) is replaced with a printed patch.

\begin{figure}[thpb]
\centering
\framebox{\parbox{3in}{
\includegraphics[width=\linewidth]{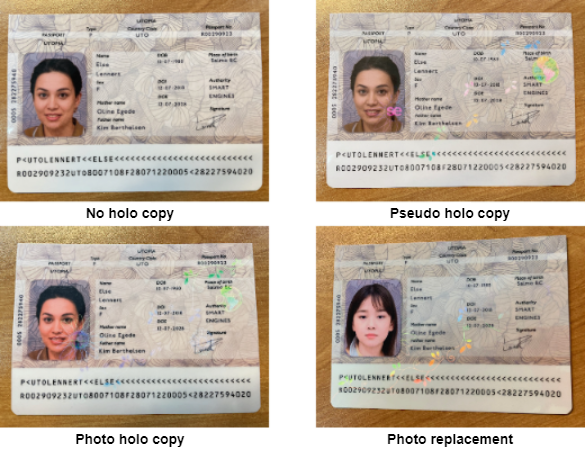}
}}
\caption{Samples of each MIDV-Holo challenges.}
\label{img:MIDVHolo_samples}
\end{figure}

It should be noted that the first three cases present challenges related to counterfeit hologram detection, while the fourth case involves a photo replacement attack, where a patch obscures part of the hologram. These forgeries result in 400 "fraud" videos, adding complexity to the detection task (see Fig.~\ref{img:MIDVHolo_samples}). They are combined with the 300 “original” videos to form the MIDVHolo dataset.

%\begin{figure}[htbp]
%\includegraphics[width=\linewidth]{Images/MIDV-holo.png}
%\caption{Samples of each MIDV-Holo challenges.} \label{img:MIDVHolo_samples}
%\end{figure}

\begin{figure}[thpb]
\centering
\framebox{\parbox{3in}{
\includegraphics[width=\linewidth]{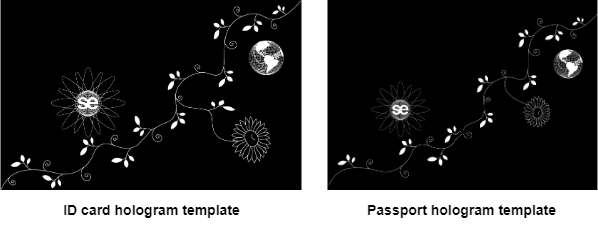}
}}
\caption{MIDV-Holo hologram templates.} 
\label{img:MIDVHolo_templates}
\end{figure}

While MIDV-Holo presents several controlled and valuable challenges, it is somewhat limited in diversity. The dataset includes only two document classes, and the hologram templates used across samples are highly similar (as illustrated in Fig.~\ref{img:MIDVHolo_templates}). Nevertheless, it remains the only publicly available dataset specifically designed for hologram detection in identity documents.

%\begin{figure}[htbp]
%\includegraphics[width=\linewidth]{Images/MIDV-holo_templates.png}
%\caption{MIDV-Holo hologram templates.} \label{img:MIDVHolo_templates}
%\end{figure}

\subsubsection{IML3 Dataset}
\begin{figure}[thpb]
\centering
\framebox{\parbox{3in}{
\includegraphics[width=\linewidth]{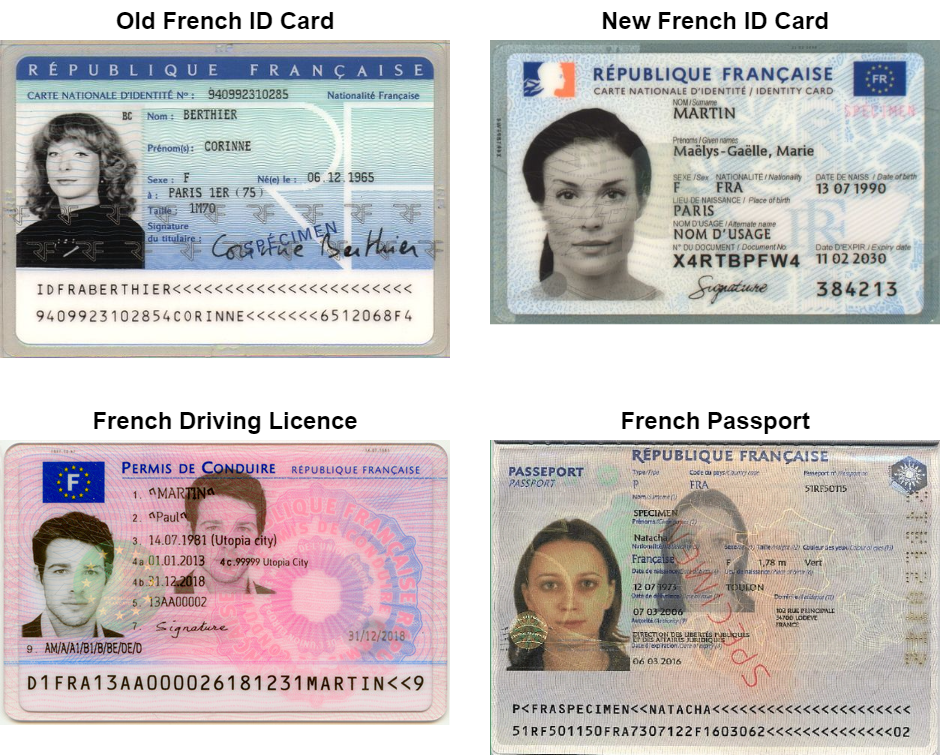}
}}
%JCB : plus court 
%\caption{All document classes represented in the IML3 dataset. These images are illustrative specimens provided by the authorities and are not part of the dataset itself. IML3 contains only real documents; the images shown here are for illustration purposes only.}
\caption{Document classes represented in IML3. These official specimen images are shown for illustration only and are not part of the IML3 dataset, which contains exclusively real documents.}
\label{img:IML3AllClasses}
\end{figure}

IML3 is a small private dataset created within the L3i laboratory, based on real French ID documents voluntarily provided by local participants. Due to the sensitive nature of the data, this dataset is not publicly available. IML3 contains a total of 748 video clips derived from 17 authentic identity documents, covering four different document types: (1) 5 old-format French ID cards, (2) 2 new-format French ID cards, (3) 6 passports, (4) 4 driving licenses. Fig.~\ref{img:IML3AllClasses} shows illustrative examples.

To simulate fake documents, each original was scanned and reprinted, producing 374 "fake" clips (i.e. half of the dataset). The videos were recorded using five different smartphones from various generations (Samsung A13, Huawei P30 Pro, Samsung Galaxy S10E, iPhone 6, Samsung Galaxy S5), across two environments with different lighting setups and brightness conditions. Clip resolutions vary depending on the smartphone (LD, HD, FHD, UHD), increasing the dataset’s realism and heterogeneity. Frames used for classification are extracted from these clips under optimal viewing conditions.

%\begin{figure}[htbp]
%\includegraphics[width=\linewidth]{Images/IML3_AllClasses.png}
%\caption{All document classes represented in the IML3 dataset. These images are illustrative specimens provided by the authorities and are not part of the dataset itself. IML3 contains only real documents; the images shown here are for illustration purposes only.}
%\label{img:IML3AllClasses}
%\end{figure}

%JCB: reformulation : plus courte
%Although IML3 includes fewer total samples than MIDV-Holo, it offers greater document diversity and aligns more closely with real-world use cases, particularly for French national IDs. However, it includes only one type of forgery (absence of a hologram), and covers only French documents, which limits the assessment of cross-national generalisation. This constitutes an acknowledged limitation of the current evaluation.
IML3 complements MIDV-Holo with greater document diversity and stronger real-world relevance, though its single forgery type and exclusive focus on French documents represent acknowledged limitations for cross-national generalisation.
\subsection{Methodology}\label{subsec:methodology}
All experiments follow a uniform evaluation protocol. The dataset is randomly split into three subsets: 70\% for training, 15\% for validation, and 15\% for testing. To avoid data leakage and ensure unbiased evaluation, all video samples from a single document are grouped together and assigned to the same subset prior to random splitting.
Each training procedure is repeated five times, and the final reported performance corresponds to the average across these five runs, ensuring greater statistical robustness.

All images are pre-processed by extracting the document from the background and resizing into 224x224. This pre-processing uses MIDV-Holo markup metadata for the images in this dataset. For IML3, the markup data are generated using SIFT descriptor and homography technique with a template, then corrected manually the few failures.

%JC: Pour court que la liste détaillé avec les equations mais on garde la définition des diminutifs (on nous le reprochera si ça n'y est pas)
To compare our method with the SotA, we report the following four metrics: False positive rate (FPR), Recall, Standard Accuracy (Acc) and F1-score (F1-S).
%\begin{itemize}
%    \item False positive rate (FPR):$\frac{FP}{Neg}$
%    \item Recall: $\frac{TP}{Pos}$
%    \item Standard Accuracy (Acc): $\frac{TP+TN}{Neg+Pos}$
%    \item F1-score (F1-S) :$\frac{2 \times Precision \times Recall}{Precision + Recall}$
%\end{itemize}
%Where TP = True Positives, FP = False Positives, TN = True Negatives, Pos = Total Positive Samples (i.e. TP+FN) and Neg = Total Negative Samples (i.e. FP+TN).

\subsection{Results and comparison}\label{subsec:results}
The results on the MIDV-Holo dataset, summarised in Table~\ref{tab:MIDV_res}, demonstrate a clear superiority of ViTransHolo over existing SotA methods. Compared to the best results of the MIDV-Holo baseline, our method achieves the following improvements: -7.33\% in FPR, +26.86\% in Recall, +17.93\% in Acc and +19.88\% in F1-S. ViTransHolo also outperforms the Triplet-loss classifier, with an additional +8.41\% in F1-S. All our metrics show significant improvements, highlighted in bold in Table~\ref{tab:MIDV_res}.

In terms of computational performance, ViTransHolo is also favorable. While it operates at a similar speed to the Triplet-loss method (or slightly slower), it is considerably faster than the MIDV-Holo baseline, which is known for its high computational cost. Concretely, inference time is in the order of seconds for the MIDV-Holo baseline versus hundreds of milliseconds for ViTransHolo and the Triplet-loss model. 
Note that the Triplet-loss baseline does not report FPR, Recall, or Accuracy, which limits the scope of direct comparison to the F1-score only. For fair comparison, we excluded the "photo replacement" fraud samples from the MIDV-Holo dataset in these experiments, as they are also omitted in the original SotA evaluations.

\begin{table}[htbp]
\caption{Results on MIDV-Holo dataset}
\begin{center}
\resizebox{\columnwidth}{!}{%
\begin{tabular}{|c|c|c|c|c|}
\hline
\textbf{Solutions / Measures} & \textbf{FPR (\%)} & \textbf{Recall (\%)} & \textbf{Acc (\%)} & \textbf{F1-S (\%)} \\
\hline
\textbf{MIDV no tracking \cite{midvholo}}&8.67&60.33&75.83&71.40\\
\hline
\textbf{MIDV tracking \cite{midvholo}}&10.33&71.33&80.50&78.53\\
\hline
\textbf{Triplet-loss \cite{pouliquen2024weakly}}&---&---&---&90\\
\hline
\textbf{ViTransHolo} &\textbf{1.34$\pm$2.90}&\textbf{98.19$\pm$1.24}&\textbf{98.43$\pm$1.28}&\textbf{98.41$\pm$1.25}\\
\hline
\end{tabular}
}%
\label{tab:MIDV_res}
\end{center}
\end{table}

\begin{table}[htbp]
\caption{ViTransHolo results on IML3 dataset}
\begin{center}
\resizebox{\columnwidth}{!}{%
\begin{tabular}{|c|c|c|c|c|}
\hline
\textbf{Exp / Measures} & \textbf{FPR (\%)} & \textbf{Recall (\%)} & \textbf{Acc (\%)} & \textbf{F1-S (\%)} \\
\hline
\textbf{IML3}&0$\pm$0&100$\pm$0&100$\pm$0&100$\pm$0\\
\hline
\textbf{IML3+MIDV-H}&1.14$\pm$1.86&99.71$\pm$0.64&99.29$\pm$1.01&99.29$\pm$0.99\\
\hline
\textbf{MIDV-Holo}&1.34$\pm$2.90&98.19$\pm$1.24&98.43$\pm$1.28&98.41$\pm$1.25\\
\hline
\end{tabular}
}%
\label{tab:IML3_res}
\end{center}
\end{table}

To validate the method in a more realistic, production-like setting, we also trained and tested ViTransHolo on the private IML3 dataset, which features real French identity documents and greater diversity in document types and hologram shapes (see Section~\ref{subsec:dataset}). Additionally, we conducted an experiment on a combined dataset (IML3 + MIDV-Holo) to verify that the model learns to detect holograms independently of document layout and content. Results of these experiments are presented in Table~\ref{tab:IML3_res}.

The results confirm that ViTransHolo remains highly effective on real-world data, including genuine French IDs. Moreover, training on the combined dataset improves generalisation and even boosts performance on the MIDV-Holo test set, likely due to increased document diversity and richer training data, which help reduce classification errors. The perfect score on IML3-only (100\% across all metrics) should be interpreted with caution given the small size of this dataset (17 source documents), which may inflate results.

\subsection{Additional experiments}\label{subsec:addExp}
To complement the previous experiments, we conducted a more detailed analysis by evaluating the model's performance on various conditions. 
These include results on each specific MIDV-Holo challenge, as well as an assessment of the network’s resilience to frugality constraints, i.e., reduced model size and limited training resources. These additional experiments aim to assess the adaptability of the model to more complex and realistic scenarios.

\subsubsection{MIDV-Holo challenges analysis}
Table~\ref{tab:MIDV_chal_res} presents the results for each specific fraud scenario within the MIDV-Holo dataset providing a fine-grained evaluation of the model.
Each sub-experiment isolates one type of challenge:
\begin{itemize}
    \item Hologram Present: Combines all original (authentic) IDs with all fraudulent samples, excluding the “photo replacement” cases. This set-up replicates the context used in previous experiments.
    \item No Hologram: Includes all “no hologram” fraud cases combined with all original IDs.
    \item Pseudo-Hologram: Consists of all pseudo-hologram fraud videos combined with original IDs.
    \item Photocopied Hologram: Includes all photocopied hologram fraud videos combined with original IDs.
    \item Photo Replacement: In this case, the model is tested on “photo replacement” frauds (considered as positive cases, because they still contain the majority of the hologram) and all other fraudulent IDs. %“Photo replacement” samples are considered positive cases (i.e. documents with a hologram), as they still contain the majority of the original hologram.%, although it is partially obscured by a replacement photo.
\end{itemize}

\begin{table}[htbp]
\caption{ViTransHolo performance on each MIDV-Holo challenge.}
%JCB : Est ce bien utile ?
%\centering\linebreak \textit{(Note: the model was not trained on photo-replacement videos; during evaluation, these samples are treated as ``Hologram Present'' cases since the original hologram remains visible.}}
\begin{center}
\resizebox{\columnwidth}{!}{%
\begin{tabular}{|c|c|c|c|c|}
\hline
\textbf{Challenge / Measures} & \textbf{FPR (\%)} & \textbf{Recall (\%)} & \textbf{Acc (\%)} & \textbf{F1-S (\%)} \\
\hline
\textbf{Hologram Present}&1.34$\pm$2.90&98.19$\pm$1.24&98.43$\pm$1.28&98.41$\pm$1.25\\
\hline
\textbf{No hologram}&1.36$\pm$1.24&94.67$\pm$8.69&97.63$\pm$1.93&95.30$\pm$4.22\\
\hline
\textbf{Pseudo-hologram}&1.36$\pm$1.24&100$\pm$0&98.98$\pm$0.93&98.04$\pm$1.77\\
\hline
\textbf{Photocopied hologram}&1.36$\pm$1.24&100$\pm$0&98.98$\pm$0.93&98.04$\pm$1.77\\
\hline
\textbf{Photo replacement}&1.78$\pm$2.90&92.78$\pm$1.26&94.07$\pm$1.52&95.98$\pm$1.03\\
\hline
\end{tabular}
}%
\label{tab:MIDV_chal_res}
\end{center}
\end{table}

Importantly, the model was not trained on any “photo replacement” samples. All such videos were reserved exclusively for testing, allowing us to evaluate the model's ability to generalise to unseen fraud types. Across all five scenarios, the model performs consistently well, with all metrics remaining above 90\%. As expected, the “photo replacement” scenario yields slightly lower scores, but the drop in performance is minimal. The relatively higher variance observed in the “no hologram” scenario (Recall: 94.67\%$\pm$8.69) may reflect greater visual heterogeneity within this fraud type, or statistical artefacts due to the small number of test videos. Further analysis would be required to confirm this hypothesis. %Nonetheless, the consistently strong performance across all scenarios further validates the model’s robustness and its ability to reliably detect the holograms under varied and realistic conditions.

\subsubsection{Reduced Train-set Study (RTS)}
To better reflect our scenario of limited sample availability, this experiment evaluates the resilience of ViTransHolo to reduced training resources. Specifically, the proportion of sample dedicated to training and validation is progressively decreased, with the aim of identifying a critical threshold beyond which the network can no longer be trained effectively.

The results are summarised in Table~\ref{tab:MIDV_RTS_res}. Between 70\% to 30\%, performance decreases at a slow rate. A clear performance drop is observed only when the training set is reduced to 30\% or less of the total dataset, which corresponds to just 180 videos instead of 420. Using only half of the original training proportion results in an accuracy loss of less than 2\%. %The trend is more clearly visible in the left curve in Figure~\ref{img:AddExpCurve}.

%JCB : pour faire plus court: 
%In conclusion, while the model cannot yet handle a true few-shot learning situation, it requires far fewer samples than expected to remain efficient. Training becomes effective with as few as 200 samples and continues to improve with more, demonstrating that even a relatively small dataset can be sufficient.
Despite not reaching few-shot learning capability, ViTransHolo proves remarkably data-efficient: training stabilises with as few as 200 samples, making it viable even in low-resource scenarios.

\begin{table}[htbp]
\caption{ViTransHolo performance in the RTS. 
%JCB :
%\centering\linebreak\textit{Note: The default model uses 70\% of the dataset for training; the 0\% setting corresponds to a pure random projection. The 'train ratio' indicates the proportion of samples used for training and validation.}}
\centering\linebreak\textit{The 'Ratio' indicates the proportion of samples used for training and validation (70\%: default setting ; 0\% representing a random baseline).}}
\begin{center}
\resizebox{\columnwidth}{!}{%
\begin{tabular}{|c|c|c|c|c|}
\hline
\textbf{Ratio} & \textbf{FPR (\%)} & \textbf{Recall (\%)} & \textbf{Acc (\%)} & \textbf{F1-S (\%)} \\
\hline
\textbf{70 - 15}&1.78$\pm$2.90&\textbf{98.64$\pm$1.24}&\textbf{98.43$\pm$1.28}&\textbf{98.43$\pm$1.25}\\
\hline
\textbf{60 - 15}&\textbf{1.60$\pm$1.12}&98.11$\pm$2.26&98.26$\pm$0.90&98.23$\pm$0.93\\
\hline
\textbf{50 - 15}&5.71$\pm$3.75&98.06$\pm$0.69&96.15$\pm$2.10&96.22$\pm$2.03\\
\hline
\textbf{40 - 10}&3.20$\pm$2.68&96.76$\pm$1.38&96.78$\pm$1.45&96.77$\pm$1.44\\
\hline
\textbf{30 - 10}&9.33$\pm$3.96&92.47$\pm$3.29&91.56$\pm$2.58&91.60$\pm$2.56\\
\hline
\textbf{20 - 5}&29.42$\pm$5.53&79.73$\pm$9.02&75.12$\pm$2.86&75.92$\pm$3.87\\
\hline
\textbf{10 - 2}&70.68$\pm$27.07&84.44$\pm$15.09&56.72$\pm$7.03&65.89$\pm$3.02\\
\hline
\textbf{0 - 0}&50.00$\pm$50.00&50.00$\pm$50.00&50.00$\pm$50.00&50.00$\pm$50.00\\
\hline
\end{tabular}
}%
\label{tab:MIDV_RTS_res}
\end{center}
\end{table}

\subsubsection{Layer Ablation Study (LAS)}
In addition to the previous experiments and to complete our frugality assessment, we conducted an ablation study of the transformer layers. By progressively reducing the number of transformer layers (the most resource-intensive components of the model), we aim to quantify the achievable gains in speed without sacrificing predictive performance. The goal is to evaluate the relationship between the number of layers, performance and computation speed.

Table~\ref{tab:MIDV_LAS_res} shows a progressive decrease in overall performance as the number of layers is reduced. This decrease remains very small (below 2\% in accuracy), while the processing time drops substantially (approximately -40\% between the 6-layers model and the 1-layer model). %To better visualise this trend, refer to the right curve in Figure~\ref{img:AddExpCurve}. 
The processing time is reported separately for CPU-only (CTime) and dual-GPU (GTime) configurations.
Based on these results, the performance loss appears minimal compared to the efficiency gains, suggesting that lighter versions of the model remain suitable in most deployment scenarios.

\begin{table}[htbp]
\caption{ViTransHolo Layer Ablation Study (LAS) results. %\centering\linebreak\textit{(Note: the default model has 6 layers; processing time is reported in milliseconds)}}
\centering\linebreak\textit{(Default model has 6 layers; processing time in milliseconds)}}
\begin{center}
\resizebox{\columnwidth}{!}{%
\begin{tabular}{|c|c|c|c|c|c|c|}
\hline
%\textbf{Layer} & \textbf{FPR (\%)} & \textbf{Recall (\%)} & \textbf{Acc (\%)} & \textbf{F1-S (\%)} & \textbf{CTime (ms)} & \textbf{GTime (ms)}\\
\textbf{Layer} & \textbf{FPR (\%)} & \textbf{Recall (\%)} & \textbf{Acc (\%)} & \textbf{F1-S (\%)} & \textbf{\begin{tabular}{@{}c@{}}CTime \\ (ms)\end{tabular}} & \textbf{\begin{tabular}{@{}c@{}}CTime \\ (ms)\end{tabular}}\\
\hline
\textbf{6}&1.78$\pm$2.90&98.64$\pm$1.24&\textbf{98.43$\pm$1.28}&\textbf{98.43$\pm$1.25}&805.4&317\\
\hline
\textbf{5}&\textbf{1.33$\pm$1.99}&97.73$\pm$1.61&98.20$\pm$1.28&98.18$\pm$1.29&739.4&247.1\\
\hline
\textbf{4}&1.78$\pm$1.86&98.18$\pm$2.96&98.20$\pm$1.00&98.17$\pm$1.06&736.2&223.9\\
\hline
\textbf{3}&3.11$\pm$2.53&\textbf{99.09$\pm$1.24}&97.98$\pm$0.94&97.99$\pm$0.91&685.2&215.3\\
\hline
\textbf{2}&4.44$\pm$3.85&98.18$\pm$2.96&96.85$\pm$2.01&96.87$\pm$1.97&626.6&207.3\\
\hline
\textbf{1}&3.11$\pm$2.53&97.73$\pm$1.61&97.30$\pm$1.51&97.29$\pm$1.50&\textbf{588.9}&\textbf{185.2}\\
\hline
\end{tabular}
}%
\label{tab:MIDV_LAS_res}
\end{center}
\end{table}

\section{CONCLUSION AND PERSPECTIVE}\label{sec:conclusion}
%JCB : Version plus concise de la conclusion :
We presented ViTransHolo, a video transformer for hologram detection within a complete remote ID verification system. Our method significantly outperforms existing SotA solutions on MIDV-Holo and generalises well to real French IDs (IML3), with further gains when both datasets are combined. Crucially, the model handles all fraud scenarios, including the unseen \textit{photo replacement} case, demonstrating strong out-of-distribution generalisation. Frugality experiments confirm that high performance is maintained with limited training data and reduced model depth, at only a marginal accuracy cost.
Looking ahead, hologram reconstruction and integrity verification constitute natural next steps, particularly verifying if the detected hologram matches the expected template: a capability that would enable more reliable identification of photo replacement forgeries.
%
%In this work, we proposed an overall system for remote identity document verification and video transformer-based approach for hologram detection in identity documents. Our hologram detection method significantly outperforms existing State-of-the-Art solutions on the public MIDV-Holo dataset, while also demonstrating strong generalisation on a private dataset of real French IDs (IML3). When trained on a combined dataset, the model achieves even better performance, benefiting from increased variability and richer training data.

%Our approach proves effective across all MIDV-Holo fraud challenges, including the difficult "photo replacement" scenario, despite the model not being explicitly trained on this type of forgery. This highlights its ability to generalise well to previously unseen cases. The additional experiments on frugality demonstrate that the model can be trained with relatively small datasets while maintaining high performance. Furthermore, it is possible to reduce the computational cost by decreasing the number of transformer layers, at the price of a slight performance drop but without loss of resilience to limited training data.
    
%With the hologram detection problem now effectively addressed, the next logical steps are hologram reconstruction and integrity verification. A particularly relevant next step is to verify whether the extracted hologram matches the expected shape thereby enabling cases such as “photo replacement” to be more reliably identified as forgeries.

%\section*{APPENDIX}

\section*{ACKNOWLEDGMENT}
This work is carried out as part of the Veridyc project (ANR-21-CE39-0003), funded by the French National Research Agency. Computing resources were provided by the L3i laboratory at La Rochelle University, funded by the French government and the Region Nouvelle-Aquitaine.
%It benefited from access to the computing resources of the L3i laboratory, operated and hosted by the University of La Rochelle. It is financed by the French government and the Region Nouvelle-Aquitaine.

%%%%%%%%%%TODO ADAPTER LES FIGURES 

\addtolength{\textheight}{-12cm}   % This command serves to balance the column lengths

%%%%%%%%%%%%%%%%%%%%%%%%%%%%%%%%%%%%%%%%%%%%%%%%%%%%%%%%%%%%%%%%%%%%%%%%%%%%%%%%

%%%%%%%%%%%%%%%%%%%%%%%%%%%%%%%%%%%%%%%%%%%%%%%%%%%%%%%%%%%%%%%%%%%%%%%%%%%%%%%%

%%%%%%%%%%%%%%%%%%%%%%%%%%%%%%%%%%%%%%%%%%%%%%%%%%%%%%%%%%%%%%%%%%%%%%%%%%%%%%%%

%%%%%%%%%%%%%%%%%%%%%%%%%%%%%%%%%%%%%%%%%%%%%%%%%%%%%%%%%%%%%%%%%%%%%%%%%%%%%%%%

%\begin{thebibliography}{99}
\bibliographystyle{splncs04}
\bibliography{Biblio.bib}
%\end{thebibliography}

\end{document}